%% file: ijcai25.tex
\documentclass{article}
\usepackage{ijcai25}

\usepackage{times}
\usepackage{soul}
\usepackage{url}
\usepackage[hidelinks]{hyperref}
\usepackage[utf8]{inputenc}
\usepackage[small]{caption}
\usepackage{graphicx}
\usepackage{amsmath}
\usepackage{amsthm}
\usepackage{booktabs}
\usepackage{algorithm}
\usepackage{algorithmic}
\usepackage[switch]{lineno}
\usepackage{amssymb}  
\usepackage{float}
\usepackage{pifont}
\usepackage{afterpage}
\usepackage{makecell}
\usepackage{multirow}
\usepackage{array}
\usepackage{adjustbox} 

\usepackage{xcolor, colortbl} 

\title{Eye-See-You: Reverse Pass-Through VR and Head Avatars}

\author{
Ankan Dash$^1$
\and
Jingyi Gu$^1$\and
Guiling Wang$^1$\and
Chen Chen$^2$
\affiliations
$^1$New Jersey Institute of Technology\\
$^2$University of Central Florida\\
\emails
\{ad892, jg95, gwang\}@njit.edu,
chen.chen@crcv.ucf.edu
}

\begin{document}

\maketitle

\begin{abstract}

   Virtual Reality (VR) headsets, while integral to the evolving digital ecosystem, present a critical challenge: the occlusion of users' eyes and portions of their faces, which hinders visual communication and may contribute to social isolation. To address this, we introduce \textbf{RevAvatar}, an innovative framework that leverages AI methodologies to enable reverse pass-through technology, fundamentally transforming VR headset design and interaction paradigms. RevAvatar integrates state-of-the-art generative models and multimodal AI techniques to reconstruct high-fidelity 2D facial images and generate accurate 3D head avatars from partially observed eye and lower-face regions. This framework represents a significant advancement in AI4Tech by enabling seamless interaction between virtual and physical environments, fostering immersive experiences such as VR meetings and social engagements. Additionally, we present \textbf{VR-Face}, a novel dataset comprising 200,000 samples designed to emulate diverse VR-specific conditions, including occlusions, lighting variations, and distortions. By addressing fundamental limitations in current VR systems, RevAvatar exemplifies the transformative synergy between AI and next-generation technologies, offering a robust platform for enhancing human connection and interaction in virtual environments.
\end{abstract}

\input{intro}

\input{relatedWorks}
\input{methods}
\input{experiments}

\input{results}

\bibliographystyle{named}
\bibliography{ijcai25}


\end{document}

%% file: intro.tex
\section{Introduction}
Augmented Reality (AR) and Virtual Reality (VR) have become critical technological advancements, transforming industries such as gaming, remote collaboration, education, and healthcare \cite{ALANSI2023100532,rambach2020surveyapplicationsaugmentedmixed,Kanschik2023}. As immersive technologies, they enable new forms of interaction and engagement, reshaping human-computer interfaces and digital experiences. 
While VR headsets have become mainstream consumer technology, they inherently isolate users from their surroundings, limiting their integration into shared environments and public spaces \cite{vrIsolation,10.1145/3290607.3299028}. Eye contact is a cornerstone of human connection and emotional communication, yet current VR headsets obscure users' eyes and facial expressions, severing visual interaction with the real world. This lack of transparency not only diminishes social presence but also leaves bystanders unaware of the user's engagement with VR content or their attentiveness. 

Addressing this fundamental limitation requires transformative AI-driven solutions to bridge the gap between virtual and physical environments. One such approach is \emph{reverse pass-through} technology, which reconstructs and displays a user’s eyes and facial expressions on the outward-facing surface of the headset. This technique enables real-time interaction, allowing bystanders to perceive eye movements and emotional expressions, effectively bridging the gap between virtual and physical environments.

\begin{figure}[!ht]
  \centering
\includegraphics[width=0.9\linewidth]{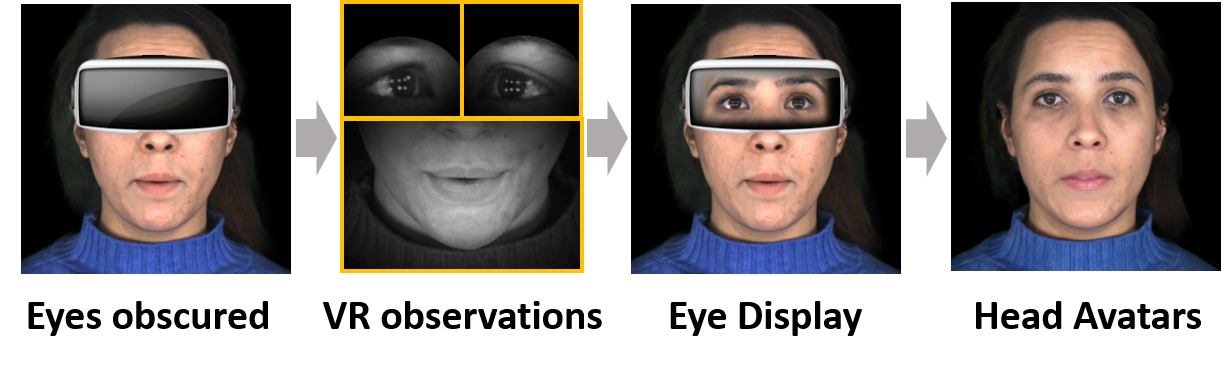}
\vspace{-0.3cm}
  \caption{Our proposed RevAvatar framework for reverse pass-through, enabling the display of eyes and full-head avatars.}
  \label{fig:teaser}
\end{figure}

Efforts to mitigate VR isolation made effort to maintain social presence, while they suffer from inauthentic eye movements \cite{10.1145/3098279.3098548,10.1145/3505270.3558348}, hardware requirements, limited facial reconstruction\cite{10.1145/3450550.3465338}, and underwhelming performance \cite{visionPro}, as shown in Table \ref{tab:related_works_comparison}.
Besides, photo-realistic avatar generation has significantly improved realism \cite{Feng:SIGGRAPH:2021,EMOCA:CVPR:2021,Zheng2023pointavatar,grassal2022neural,li2023goha}, but limited by the need for multi-view images or specialized VR headsets. Person-specific avatar generation struggles with high customization needed, hindering widespread adoption \cite{10.1145/3197517.3201401,10.1145/3306346.3323030,10.1145/3386569.3392493}. 

\input{tables/existing_works}

To address the limitations of current VR systems, we propose RevAvatar, a framework to generalize across VR headsets with minimal device-specific fine-tuning. Unlike existing solutions, \textbf{RevAvatar} leverages advanced AI techniques to overcome the isolation caused by VR headsets by reconstructing and displaying the user’s eyes and facial expressions in real-time, enabling seamless interaction between virtual and physical environments. Additionally, RevAvatar facilitates the creation of full-head 3D avatars, enhancing immersive experiences for applications like virtual meetings. The main pipeline combines real-time 2D face restoration for "reverse pass-through" and a one-shot 3D avatar generation model, achieving 0.008-second inference on mobile SoCs like Apple M2. It is compatible with consumer mixed-reality devices such as the Apple Vision Pro and upcoming Samsung and Google VR headsets. Crucially, it eliminates the need for 3D scans, requiring only a selfie-like Digital Persona (DP) image, improving accessibility and convenience.

Developing generalized solutions for diverse VR headsets is challenging due to variability in camera specifications and placements across brands like Apple, Samsung, Meta, Vive, and Varjo. To address this, we introduce \textbf{VR-Face}, a novel dataset comprising 200,000 samples that simulate diverse VR conditions, including occlusions, lighting variations, and distortions. VR-Face not only supports RevAvatar’s development but also provides a foundational resource for advancing research in AI-driven VR technologies.
\paragraph{Our contributions are:} \ding{172} \textbf{RevAvatar Framework}: We introduce \textbf{RevAvatar}, an AI-driven framework for real-time reverse pass-through and 3D avatar generation in VR. This solution enhances VR immersion by eliminating the need for user-specific models or custom hardware.
\ding{173} \textbf{Efficient AI for Mobile SoCs}: Our 2D face reconstruction model operates efficiently on mobile SoCs, such as the Apple M2 in Apple Vision Pro, achieving an inference time of just 0.008 seconds. This demonstrates its scalability across diverse VR devices.
\ding{174} \textbf{VR-Face Dataset}: We present \textbf{VR-Face}, a public dataset with 200,000 samples simulating challenging VR scenarios. It supports the development of AI technologies adaptable to various headset specifications and advances research in VR.
\ding{175} \textbf{AI-Enabled VR Advancements}: Through RevAvatar and VR-Face, we drive significant AI innovations that address key VR challenges like user isolation and hardware diversity, setting a new standard for AI-driven progress in VR.

%% file: tables/existing_works.tex
\begin{table*}[h]
    \centering
    \fontsize{9}{9}\selectfont
    \renewcommand{\arraystretch}{0.9} 
    \begin{tabular}{p{1.5cm} p{3.5cm} p{3.5cm} p{7.5cm}} 
        \toprule
        & \textbf{Methods} & \textbf{Pros} & \textbf{Cons} \\
        \midrule
        \multirow{4}{*}{Reverse Pass} 
        & FrontFace, Google Eyes & Displays animated eyes & Lacks authenticity; fails to convey emotions \\
        & \cite{10.1145/3450550.3465338} & Improves social presence  & Requires custom hardware; lacks full facial reconstruction \\
        & EyeSight-Apple Vision Pro & Displays eyes externally & Limited realism; unclear effectiveness \\
        & \textbf{Ours} &  Full face reconstruction & - \\
        \midrule
        \multirow{3}{*}{Avatar} 
        & Photo-realistic Avatar & Increases realism & Requires multi-view images and specialized VR headsets \\
        & Person-specific Avatar & Enables personalization & High customization limits scalability \\
        & \textbf{Ours} & One-shot avatar & - \\
        \bottomrule
    \end{tabular}
    \caption{Comparison of Existing VR Social Presence Methods}
    \label{tab:related_works_comparison}
\end{table*}

%% file: relatedWorks.tex
\section{Related Work}
\paragraph{Eye tracking based animation}
FrontFace \cite{10.1145/3098279.3098548} and Googly Eyes \cite{10.1145/3505270.3558348} aimed to represent user attention and gaze using animated eye movements via eye tracking and Head-Mounted Displays (HMDs). However, these approaches focus on displaying {\sl animated} eye states, such as whether the eyes are open or closed, and the gaze direction, without conveying genuine emotions or expressions.

\paragraph{Eye and Face Reconstruction: Reverse Pass-Through}
A ``reverse pass-through" prototype headset \cite{10.1145/3450550.3465338} reconstructs and displays users' eyes on an external screen but requires costly and custom hardware inaccessible to most users.
Apple's Vision Pro with ``EyeSight" projects eyes onto an external display, but its functionality remains unclear \cite{visionPro}, and early reviews suggest underwhelming performance \cite{visionProReview,visionProReview2}.
Other face-restoration methods from partial VR headset data often require multiple views or customized headsets, making them impractical for widespread use. They also rely on user-specific avatars, requiring costly individualized training and limiting real-world use \cite{10.1145/3197517.3201401,10.1145/3306346.3323030,10.1145/3386569.3392493}.

\paragraph{Face image composition and one-shot Avatar generation}
Several methods were proposed for face composition and synthesis.
PixelStyle2Pixel (PSP) \cite{psp} performs image-to-image translation using StyleGAN’s latent space, while failing to preserve the identity of unseen individuals. StyleMapGAN \cite{kim2021stylemapgan} faces similar identity preservation challenges during inference. 
One-shot photo-realistic avatar generation made significant strides, like ROME \cite{Khakhulin2022ROME} generating mesh-based avatars from single images, and CVTHead \cite{ma2023cvthead} using transformers and point-based neural rendering. Portrait4D \cite{deng2024portrait4d} employs a part-wise 4D generative model for synthesizing multi-view images and leverages transformers to create highly detailed, animatable avatars. They are closest to our task of one-shot avatar generation.

\paragraph{VR simulation dataset}
Despite growing interest in VR simulations, publicly available datasets remain scarce. Eye images captured by IR cameras in VR headsets suffer from occlusions and limited fields of view. MEAD data \cite{showYourFace} was used for VR simulations but lacked real-world scenario complexity, particularly in occlusions like eyebrow obstruction. Other works \cite{10.1145/3197517.3201401,10.1145/3306346.3323030} used custom VR headsets with IR cameras for optimal eye capture, but these setups are not generalizable to commercial headsets, and datasets are publicly inaccessible.

%% file: methods.tex
\input{dataset}

\section{Methodology}
\begin{figure}[!ht]
  \centering
  \includegraphics[width=0.9\linewidth]{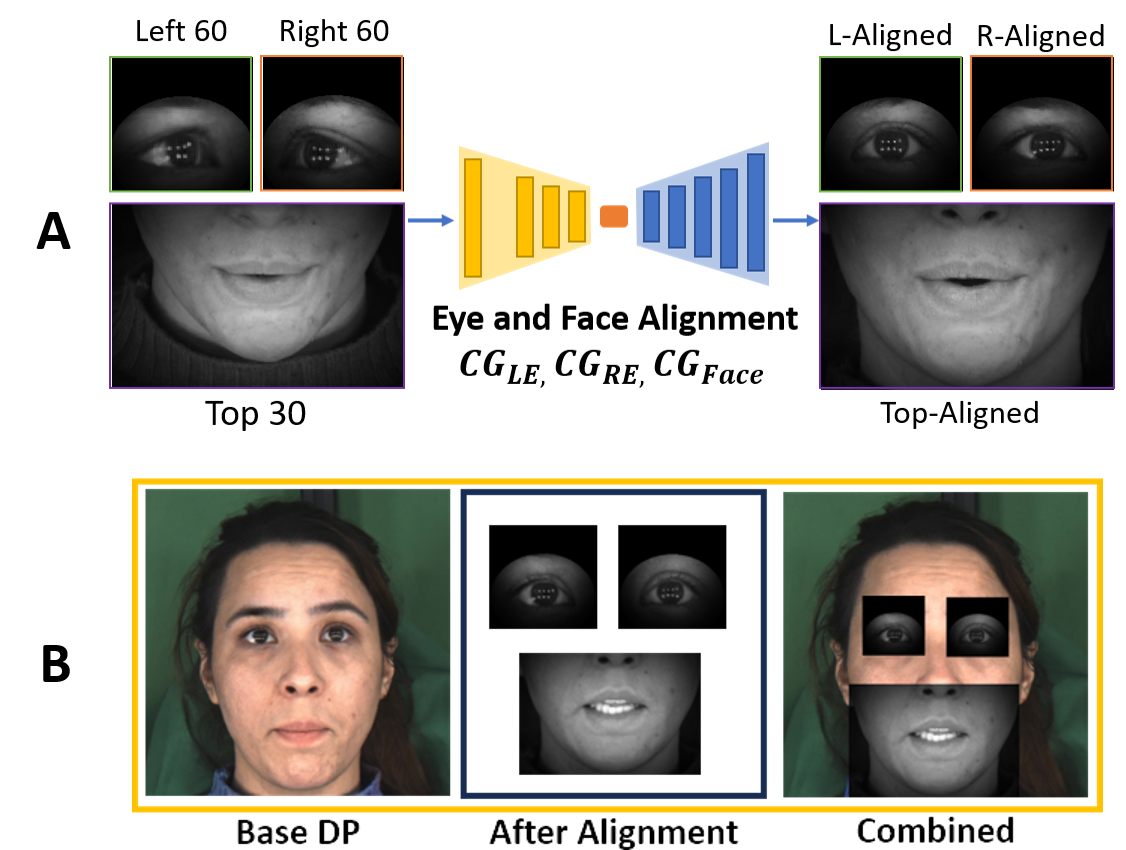}
  \caption{A - Left, Right Eye and Face alignment models based on CycleGAN. B - The combined image is used as input for full face restoration.}
  \label{fig:alignment}
\end{figure}

\emph{RevAvatar} comprises an initial setup stage and the main pipeline, which consists of image alignment, 2D face restoration, and 3D avatar generation for reverse pass-through and immersive VR experience, as shown in Figure \ref{fig:alignment}. and \ref{fig:framework}.

During the initial VR headset setup, the user is prompted to capture a selfie using the headset's external camera or upload an image via their online account. Manufacturers provide detailed instructions to ensure the image captures the entire face without occlusions and under good lighting. The image is then processed using a facial landmark model\cite{bulat2017far} for cropping and alignment in subsequent stages. This selfie, referred to as the Digital Persona (DP) image, serves as the reference for 2D and 3D reconstruction.

The main pipeline aligns or frontalizes the processed eye and face-tracking images. A lightweight GAN-based model\cite{NIPS2014_5ca3e9b1} is then used for full face restoration. For Avatar generation, we combine 3DMM\cite{egger20203dmorphablefacemodels} models with neural networks for tri-plane representation and volume rendering to achieve accurate 3D reconstruction.

\subsection{Eye and Face Alignment}
\label{subsec:Eye and Face Alignment}
We use VR-Face dataset, consisting of non-aligned images of left and right eyes, with the goal of aligning or frontalizing these images (Figure \ref{fig:alignment}-A). To achieve this, we adopt a CycleGAN-based framework \cite{cycleGAN}, which learns bidirectional mappings between tilted eye images (domain $A$) and frontal eye images (domain $B$) through Cycle-Consistency. We use two distinct alignment models, $CG_{LE}$ and $CG_{RE}$ for each eye, and use a separate model $CG_{Face}$ to frontalize the lower face image. To preserve gaze, we use eye landmark detection \cite{bulat2017far} and compute Gaze Estimation Error as the angular difference between aligned and ground truth images, achieving errors below one degree. Aligned eye and lower face images are then pasted onto the DP image using facial landmarks. Dataset details are in Section \ref{sec:data}, with examples in Figure \ref{fig:alignment}-B.

\input{figures/overview}

\subsection{2D Full Face Restoration}

To restore a full face image from frontalized eye and face tracking images, we develop a lightweight GAN-based restoration model with a Generator $G$ and Discriminator $D$. It takes aligned grayscale left and right eye images, along with a lower face image pasted onto the color DP image, and generates a restored output where the grayscale images blend seamlessly with the rest of the face, capturing facial expressions. Figure \ref{fig:framework}. shows our 2D Face Restoration Framework. To handle high occlusion of input eye images, the model uses a reference image for reconstructing facial features, providing context for occluded areas like eyebrows. During training, random reference images from various users are selected to improve robustness. During deployment, the DP image serves as a reference for reconstructing occluded areas.
 
The Generator \( G \) comprises an Input Encoder \( E_I \) and a Reference Encoder \( E_R \), which share a similar structure. \( E_I \) takes the DP image with partial VR observations pasted on top of the base DP image \( x \) as the input, and the reference DP image \( z \) is given as input to \( E_R \). To enhance the model’s ability to capture both global and fine-grained details, the encoders are extended to operate at multiple scales, where each scale extracts features at different levels of resolution. The Generator is built on multiple ResNet blocks, with each block processing features at different resolutions. We use cross-attention module to align and integrate features from the reference image with those of the input image, enhancing context-aware and guided image generation. The cross-attention module operates at multiple scales, where each scale learns to focus on different levels of detail, from coarse structures to fine textures. This multi-scale feature fusion ensures the generation of high-fidelity images by combining both global context and local detail. The architecture begins with initial convolutional layers for downsampling through \( E_I \) and \( E_R \), extracting multi-scale features. The multi-scale features are then processed through cross-attention and residual block processing. Finally, the decoder reconstructs the synthesized image \( G(x, z) \), utilizing the multi-scale features to improve the quality of the generated image at all levels. We use a Multiscale Discriminator, which employs multiple instances of a PatchGAN \cite{8100115} discriminator, each responsible for evaluating the image at a specific scale.

The Generator \( G \) is trained to minimize the following loss functions. (1) The adversarial loss \( \mathcal{L}_{adv} \) ensures that the generated image \( G(x, z) \) is indistinguishable from real images \( y \) by the Discriminator \( D \). (2) The L1 loss \( \mathcal{L}_{L1} \) ensures the generated image \( G(x, z) \) is close to the ground truth image \( y \). (3) The LPIPS (Learned Perceptual Image Patch Similarity) \cite{zhang2018perceptual} loss \( \mathcal{L}_{LPIPS} \) assesses perceptual similarity between the generated image \( G(x, z) \) and the target image \( y \). The total loss is defined as $ \mathcal{L}_{\text{total}} = \mathcal{L}_{\text{G}} + \mathcal{L}_{\text{D}} $.

\vspace{-0.5cm}

\begin{align}
   \mathcal{L}_{G} = \lambda_{adv} \mathcal{L}_{adv} + \lambda_{L1} \mathcal{L}_{L1} + \lambda_{LPIPS} \mathcal{L}_{LPIPS} 
\end{align}

\vspace{-0.5cm}

\begin{align}
\mathcal{L}_{adv} = \mathbb{E}_{y} \left[ \log D(y) \right] + \mathbb{E}_{x} \left[ \log(1 - D(G(x, z))) \right]
\end{align}
where the weights \( \lambda_{adv} \), \( \lambda_{L1} \), and \( \lambda_{LPIPS} \) balance the contributions of each loss term.

The Discriminator \( D \) is trained to distinguish between real images and generated images using the adversarial loss:
\begin{align}
\mathcal{L}_{D} = \mathbb{E}_{y} \left[ \log D(y) \right] + \mathbb{E}_{x} \left[ \log(1 - D(G(x, z))) \right]
\end{align}
where \( D(y) \) is the probability that the image \( y \) is real, and \( D(G(x, z)) \) is the probability that the generated image \( G(x, z) \) is real.

\subsection{One-shot 3D Head Avatar Model}

We extend our reverse pass-through system to generate full-head avatars for immersive VR, building on recent one-shot facial avatar generation advancements. This approach overcomes the limitations of requiring specialized models for each subject or multi-view inputs, addressing challenges that hinder practical real-world applications.

Our Framework uses the user's DP or selfie image as the source image $I_{s}$ and the reconstructed image from 2D full-face restoration serves as the driving or target image $I_{t}$. The source image is used to extract the identity, and the target image is responsible for providing the pose and expression information. Our framework comprises three main branches which include the Global Branch $E_{G}$, Detail Branch $E_{D}$ and the Expression Branch $E_{E}$. The output is then up-scaled and refined using a super-resolution module. 

The $E_{G}$ branch uses a hybrid transformer model with a series of convolutional and transformer blocks along with SegFormer\cite{xie2021segformersimpleefficientdesign} to generate a tri-plane representation. We use SegFormer as it allows for effective mapping from 2D space to 3D space. This is achieved by predicting a tri-plane \(T_g\) that represents the neutral expression of the human face in a canonical 3D space. To ensure that the generated tri-plane \(T_g\) aligns with the identity of \(I_{s}\) and maintains a neutral expression, we incorporate a 3D Morphable Model (3DMM) to render a face with the same identity and camera pose as the source image, but with a neutral expression. The Detail Branch $E_{D}$ builds on the  geometry provided by the Global Branch by capturing and reconstructing intricate facial details from the source image \(I_{s}\).  The features of the Detail Branch are transferred to the global triplane, creating an appearance triplane \(T_{d}\) that improves the initial reconstruction with fine-grained details, such as texture and surface features. The Expression Branch focuses on modeling and transferring the expression from the target image \(I_t\) onto the reconstructed 3D avatar. This branch utilizes a 3DMM to predict the shape and expression coefficients for both the source image \(I_s\) and target image \(I_t\). The expression coefficients of \(I_t\) are used to render a frontal-view expression image \(I_e\), which is then encoded into an expression tri-plane \(T_{e}\). This expression tri-plane is added to the canonical tri-plane \(T_{g}\) along with the appearance tri-plane \(T_d\) to generate the final 3D reconstruction. The integration of these three branches allows the model to combine the identity from the source image with the expression and head pose from the target image, effectively transferring the desired expression onto the source image while maintaining high fidelity in both appearance and geometry. Given the high computational demands of volumetric rendering, we first render low-resolution images and then use a super-resolution\cite{wang2021gfpgan} module to produce the final high-quality output.

We use a two-stage training schedule for multi-view consistency and efficiency. In the first stage, the model trains at a lower resolution without an upscaling module, optimizing $L_1$ and $L_{LPIPS}$ losses between the Global Branch feature rendering and the 3DMM rendering of the source image, as well as the combined tri-plane features and target image.
\vspace{-0.1cm}
\begin{align}
\mathcal{L}_{\text{G}} &= L_1(R(T_{\text{g}}), R_{\text{3DMM}}(I_{\text{s}})) \nonumber \\
&\quad + L_{\text{LPIPS}}(R(T_{\text{g}}), R_{\text{3DMM}}(I_{\text{s}})) \\
\mathcal{L}_{\text{Combined}} &= L_1(R(T_{\text{combined}}), I_{\text{t}}) \nonumber \\
&\quad + L_{\text{LPIPS}}(R(T_{\text{combined}}), I_{\text{t}}) \\
\mathcal{L}_{\text{Stage1}} &= \lambda_{\text{G}} \mathcal{L}_{\text{G}} + \lambda_{\text{Combined}} \mathcal{L}_{\text{Combined}}
\end{align}
\vspace{-0.1cm}
In the second stage of training, we only fine-tune the upscaling module using $L_1$, $L_{LPIPS}$, and GAN loss objective. Additionally, we use an eye region loss which calculates the $L_1$ between only the eye region rendering the output image and the target ground truth image to ensure accurate gaze.
\vspace{-0.1cm}
\begin{align}
\mathcal{L}_{\text{Stage2}} &= \lambda_{\text{L1}} L_1(I_{\text{o}}, I_{\text{t}}) + \lambda_{\text{LPIPS}} L_{\text{LPIPS}}(I_{\text{o}}, I_{\text{t}}) \nonumber \\
&\quad + \lambda_{\text{GAN}} \mathcal{L}_{\text{GAN}} + \lambda_{\text{Eye}} L_1(I_{\text{eye-output}}, I_{\text{eye-target}}) \nonumber \\
&\quad + L_{\text{LPIPS}}(I_{\text{eye-output}}, I_{\text{eye-target}}) \\
\mathcal{L}_{\text{Total}} &= \mathcal{L}_{\text{Stage1}} + \mathcal{L}_{\text{Stage2}}
\end{align}

\subsection{Reverse Pass-through and Avatar Outputs}
\begin{figure}[!ht]
  \centering
\includegraphics[width=0.9\linewidth]{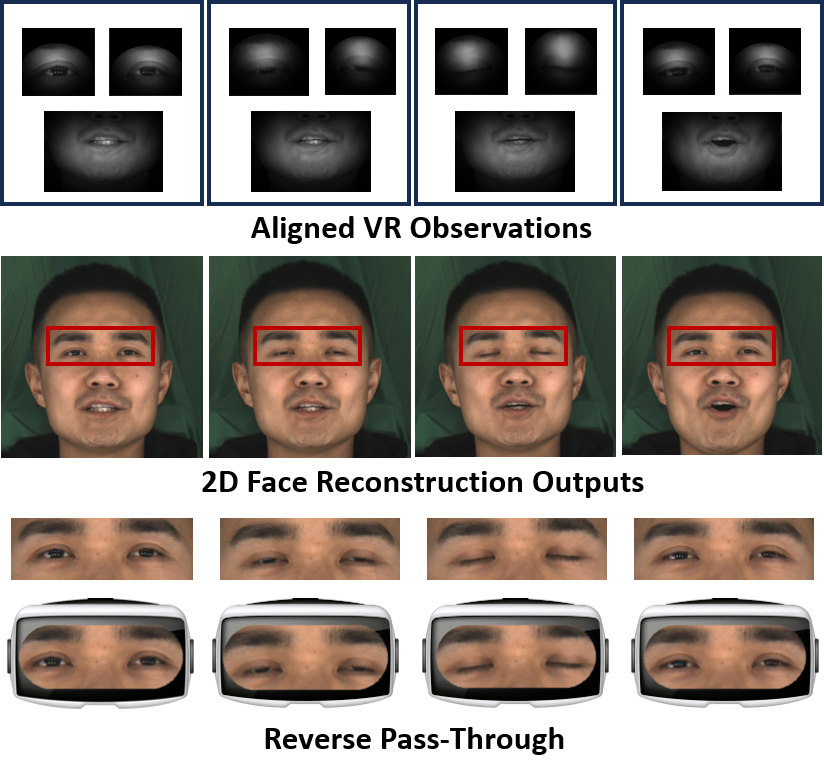}
  \caption{Sample output from 2D Face Restoration model which can be used for Reverse pass-through.}
  \label{fig:reversePass}
\end{figure}

The output from our Full Face Restoration model can be cropped to display the eye region and realize reverse pass-through on mainstream VR headsets(Figure \ref{fig:reversePass}). This allows users to maintain eye contact and convey expressions. Additionally, the outputs from our head avatar model can be leveraged for immersive applications such as VR meetings and the metaverse, providing visually accurate and expressive 3D avatars that enhance virtual interactions and communication.

%% file: dataset.tex
\section{VR-Face Dataset}
\begin{figure}[!ht]
  \centering
  \includegraphics[width=0.9\linewidth]{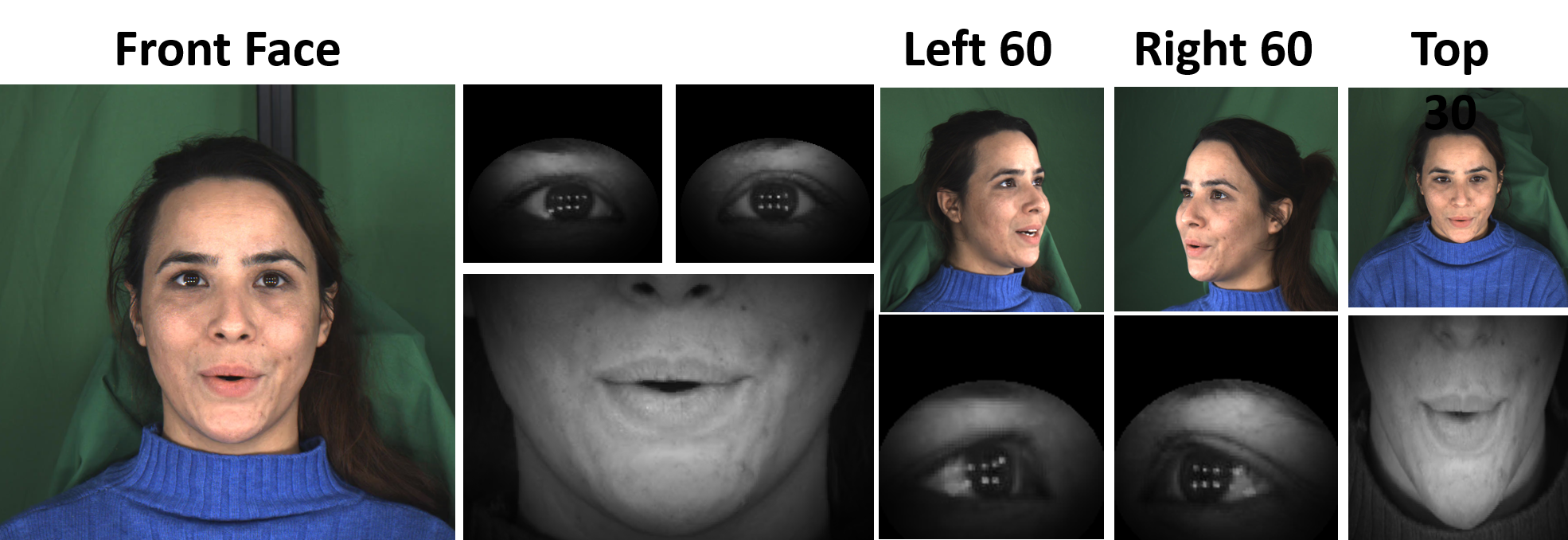}
  \caption{Sample processed images from the \emph{VR-Face} dataset simulating VR environments.}
  \label{fig:dataset}
\end{figure} 

\begin{table}[h]
    \centering
    \fontsize{9}{9}\selectfont
    \begin{tabular}{p{2cm} p{5.5cm}} 
        \toprule
        \textbf{Aspect} & \textbf{Description} \\
        \midrule
        Total Samples & 200,000 \\
        Image Types & Full-face, left eye, right eye, lower face \\
        Angles & Front, left-60°, right-60°, top-30° \\
        Expressions & Anger, contempt, disgust, fear, happy, neutral, sad, surprise \\
        Preprocessing & Distortion, masking, vignetting, blur, grayscale \\
        Effects & Occlusions, lighting shifts, noise, eyebrow reduction \\
        Diversity & Skin tones, genders, ethnicities \\
        \bottomrule
    \end{tabular}
    \caption{Overview of the VR-Face Dataset}
    \label{tab:vr_face_dataset}
\end{table}

We develop the \emph{VR-Face} dataset (Figure \ref{fig:dataset}), containing 200,000 samples, each comprising a full-face image, left and right eye images from various angles, and a lower-face image. The dataset captures a wide range of facial expressions and perspectives, with pre-processing to simulate visual effects observed in VR headset imagery (details in Table \ref{tab:vr_face_dataset}). VR-Face is designed to be inclusive, representing diverse skin tones, sex, race, and ethnicity. While it serves as a benchmark for reverse pass-through and analysis, it is not intended to replace real-world VR headset data but can be adapted to specific devices with suitable datasets.

%% file: figures/overview.tex
\begin{figure*}[!ht]
  \centering
\includegraphics[width=0.9\linewidth]{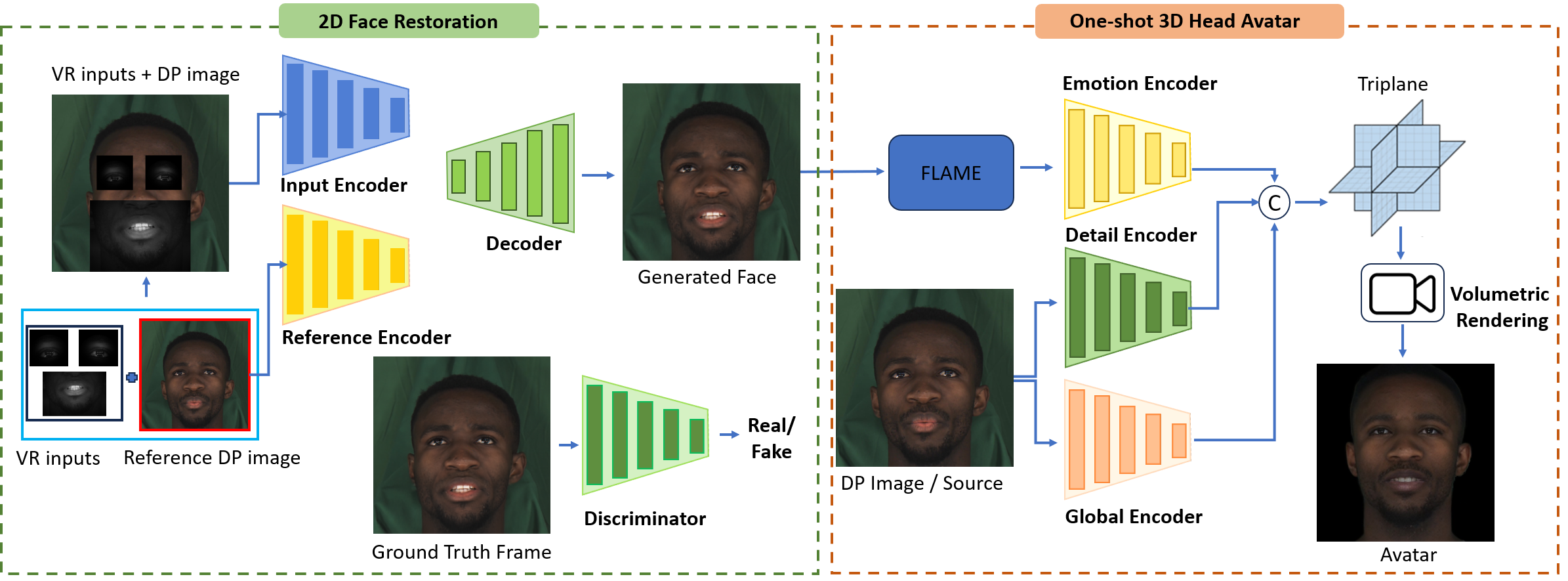}
  \caption{Overview of the \textbf{2D Face Restoration} and \textbf{One-Shot Avatar} generation model. \textbf{2D Face Restoration:} Partial VR observations and the reference DP are inputs to the Input Encoder, while the reference image is processed by the Reference Encoder. The restored face output drives the one-shot avatars. \textbf{One-Shot Avatar:} The DP image serves as the source, and the restored image from the 2D face restoration model drives the avatar generation. A tri-plane is generated from concatenated encoder outputs, followed by volumetric rendering and super-resolution to produce the final output.}
  \label{fig:framework}
\end{figure*}

%% file: experiments.tex
\input{tables/reconstruction}

\section{Experiments}
\subsection{Datasets}\label{sec:data}
We utilize \emph{VR-Face} as the main dataset to train and test our framework, and three additional datasets for training to enhance the generalization. (1) The Eye and Face Alignment model is exclusively trained and tested on \emph{VR-Face} dataset. (2) For 2D face restoration model, we integrate CelebHQ\cite{celebAHQ} and FFHQ\cite{ffhq} datasets, which contain images only, in conjunction with \emph{VR-Face} dataset for training, allowing for better generalization across different skin tones and facial attributes. (3) For 3D avatar generation, besides \emph{VR-Face}, we further leverage FFHQ, CelebV-HQ\cite{zhu2022celebvhq}, VFHQ\cite{vfhq} datasets to provide a rich set of facial attributes and emotional variations.

\begin{figure*}[!ht]
  \centering
  \includegraphics[width=0.9\linewidth]{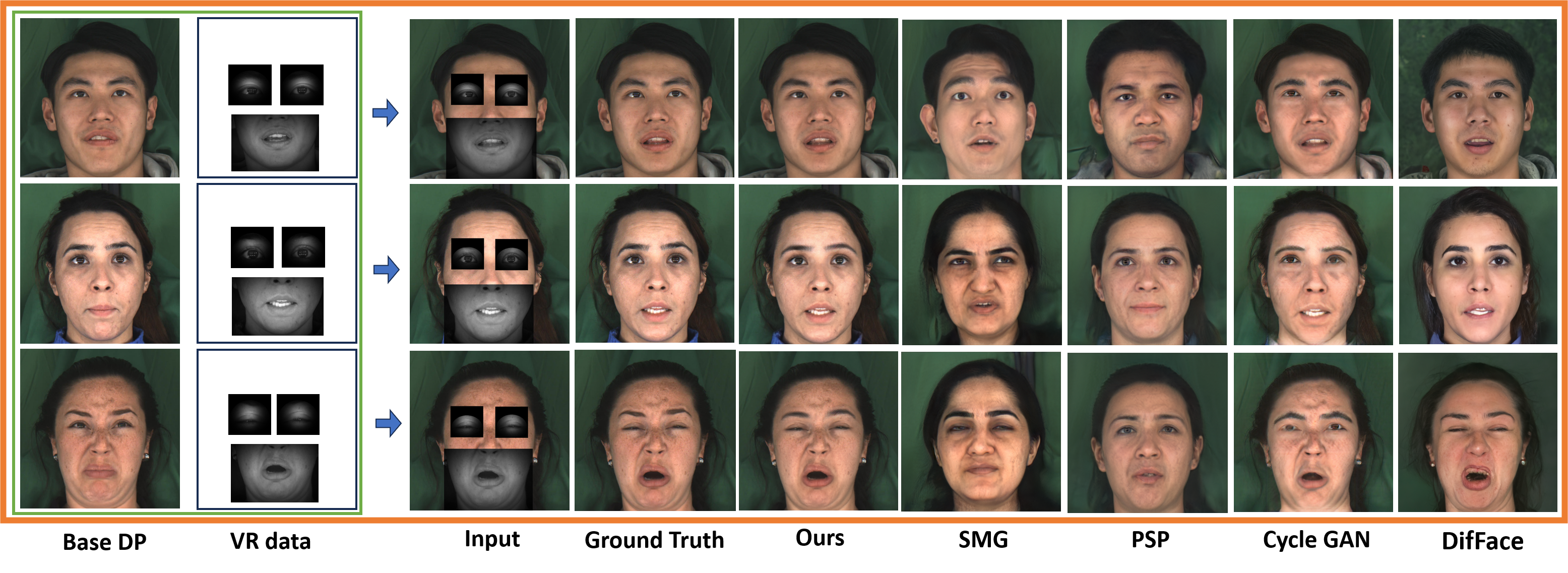}
  \caption{Qualitative comparison of full-face restoration results on unseen test data.}
  \label{fig:FaceRestorationResultsDiffRes}
\end{figure*}

\subsection{Implementation}
All models are trained on $512 \times 512$ images. The Alignment and 2D face restoration models use a single A100 GPU, while the 3D head avatar model is trained on 4 A100 GPUs. We train three alignment models: two for the eyes and one for the lower face. Batch sizes are set to 32 for Alignment, 16 for 2D face restoration, and 4 for the 3D avatar model. All models use the Adam optimizer with a 0.0001 learning rate. During inference, the alignment models run in 0.004s, the face restoration model in 0.006s per image, and the avatar model achieves 22 FPS. All inferences are performed on a single A100 GPU including mobile GPUs such as Apple M2 GTX 1050, and MX350.

\input{tables/inference_time}

\subsection{Baseline}
Given the lack of well-established baseline models for reverse pass-through VR, a direct holistic comparison of our entire framework is not feasible. Instead, we evaluate our 2D full-face restoration model by comparing it with state-of-the-art GAN-based approaches in image composition and reconstruction: CycleGAN, PSP \cite{psp}, and SMG \cite{kim2021stylemapgan}. Additionally, we include DifFace \cite{10607954} as a diffusion-based baseline. Although we considered diffusion models such as OSDFace \cite{wang2024osdfaceonestepdiffusionmodel}, DifFace \cite{10607954}, OSEDiff \cite{wu2024onestepeffectivediffusionnetwork}, and DiffBIR \cite{lin2024diffbir}, \textbf{their inference times of 0.1, 6.1, 0.12, and 8.01 seconds, respectively, on an A100 GPU make them unsuitable for real-time applications}. In contrast, CycleGAN, PSP, SMG, and our model achieve real-time performance with \textbf{inference times of 0.016, 0.041, 0.038, and 0.006 seconds, respectively}. For 3D avatar reconstruction, we benchmark our model against leading one-shot approaches: ROME \cite{Khakhulin2022ROME}, CVTHead \cite{ma2023cvthead}, and Portrait-4D \cite{deng2024portrait4d}.

%% file: tables/reconstruction.tex
\begin{table*}
    \centering\fontsize{9}{9}\selectfont
    \begin{tabular}{cccccccc}
        \toprule
        \multirow{2}*{\textbf{Model}} & \multicolumn{3}{c}{\textbf{Full Face}} & \multicolumn{3}{c}{\textbf{Eye Region of Interest}} & \multirow{2}*{\textbf{Inference Time (s)}} \\ 
        \cline{2-7} 
         & \textbf{SSIM}$\uparrow$ & \textbf{PSNR}$\uparrow$ & \textbf{LPIPS}$\downarrow$ & \textbf{SSIM}$\uparrow$ & \textbf{PSNR}$\uparrow$ & \textbf{LPIPS}$\downarrow$ & \\ 
        \midrule
        
        CycleGAN   & 0.8414 & 23.0429 & 0.0618 & 0.6711 & 20.2291 & 0.1122 & 0.016 \\
        PSP        & 0.6271 & 19.6809 & 0.1714 & 0.5737 & 18.7591 & 0.1694 & 0.041 \\
        SMG        & 0.6521 & 21.0219 & 0.2349 & 0.6211 & 19.2129 & 0.1521 & 0.039 \\
        DifFace (2024)    & \textbf{0.9541} & 29.8129 & 0.1306 & 0.8122 & 26.0021 & 0.1023 & 6.125 \\
        Ours       & 0.9445 & \textbf{31.3951} & \textbf{0.0243} & \textbf{0.8572} & \textbf{28.2897} & \textbf{0.0510} & \textbf{0.006} \\
       
        \bottomrule
    \end{tabular}
    \caption{Quantitative Comparison for Full Face and Eye Region Reconstruction, including Inference Time (s).}
    \label{tab:face_eyes_reconstruction_with_inference}
\end{table*}

%% file: tables/inference_time.tex
\begin{table}[!ht]
    \centering\fontsize{9}{9}\selectfont
    \begin{tabular}{ccccc}
        \toprule
        \multirow{2}{*}{\textbf{Model}} & \multicolumn{4}{c}{\textbf{Inference Time (s)}}  \\ 
        \cline{2-5} 
        & \textbf{A100} & \textbf{Apple M2} & \textbf{MX350} & \textbf{GTX 1050} \\ 
        \midrule
        
        CycleGAN   & 0.016 & 0.020 & 0.350 & 0.278 \\
        PSP        & 0.041 & 0.050 & 0.852 & 0.544 \\
        SMG         & 0.039 & 0.045 & 0.791 & 0.365 \\
        Ours       & \textbf{0.006} & \textbf{0.012} & \textbf{0.125} & \textbf{0.0517} \\
       
        \bottomrule
    \end{tabular}
    \caption{Inference Time for A100, Apple M2, MX350, and GTX 1050 for Full Face  Reconstruction for Reverse Pass-Through.}
    \label{tab:face_eyes_reconstruction_inference_m2_a100_mx350}
\end{table}

%% file: results.tex
\subsection{Results}

For our 2D face reconstruction model and avatar, we generate images at 512x512 resolution and evaluate them against ground truth images using three metrics: SSIM \cite{ssim}, PSNR, and LPIPS. These metrics assess visual accuracy and perceptual quality by considering structural similarity, pixel-level differences, and perceptual relevance.

\paragraph{2D Face Restoration}
Figure \ref{fig:FaceRestorationResultsDiffRes}. and Table \ref{tab:face_eyes_reconstruction_with_inference}. presents the qualitative and quantitative comparison between our face restoration model and baselines. Despite its lightweight design, our model performs better than other baselines. While CycleGAN performs comparably to other GAN-based models, it struggles to effectively colorize and blend eye and lower face, leading to severe artifacts, as shown in Figure \ref{fig:FaceRestorationResultsDiffRes}. PSP and SMG, which rely on StyleGAN-based generators, map inputs to a latent space, resulting in a loss of identity and inaccurate face restorations. During testing, SMG tends to output similar images from its training set but with altered expressions, while PSP produces outputs that often diverge significantly from the ground truth, as highlighted in Figure \ref{fig:FaceRestorationResultsDiffRes}. This exposes a critical limitation of StyleGAN-based models: poor generalization on unseen data. DifFace, as a diffusion-based model, achieves high SSIM and PSNR for full-face reconstruction, outperforming other baselines in preserving global facial structure. However, it struggles to retain individual identity, leading to subtle yet noticeable shifts in facial features. Additionally, the iterative nature of the diffusion process results in significantly higher inference time, making DifFace less suitable for real-time applications.
In contrast, our model excels at handling unseen face images, demonstrating superior generalization capabilities. It achieves the highest PSNR and the lowest LPIPS, indicating better perceptual quality and sharpness. Moreover, its inference time is orders of magnitude faster than DifFace, making it highly efficient for real-time applications. Figure \ref{fig:reversePass} depicts how the output of our face restoration model enables real-time reverse pass-through capabilities in VR applications.

\input{tables/avatar}

\begin{figure}[!ht]
  \centering
  \includegraphics[width=0.9\linewidth]{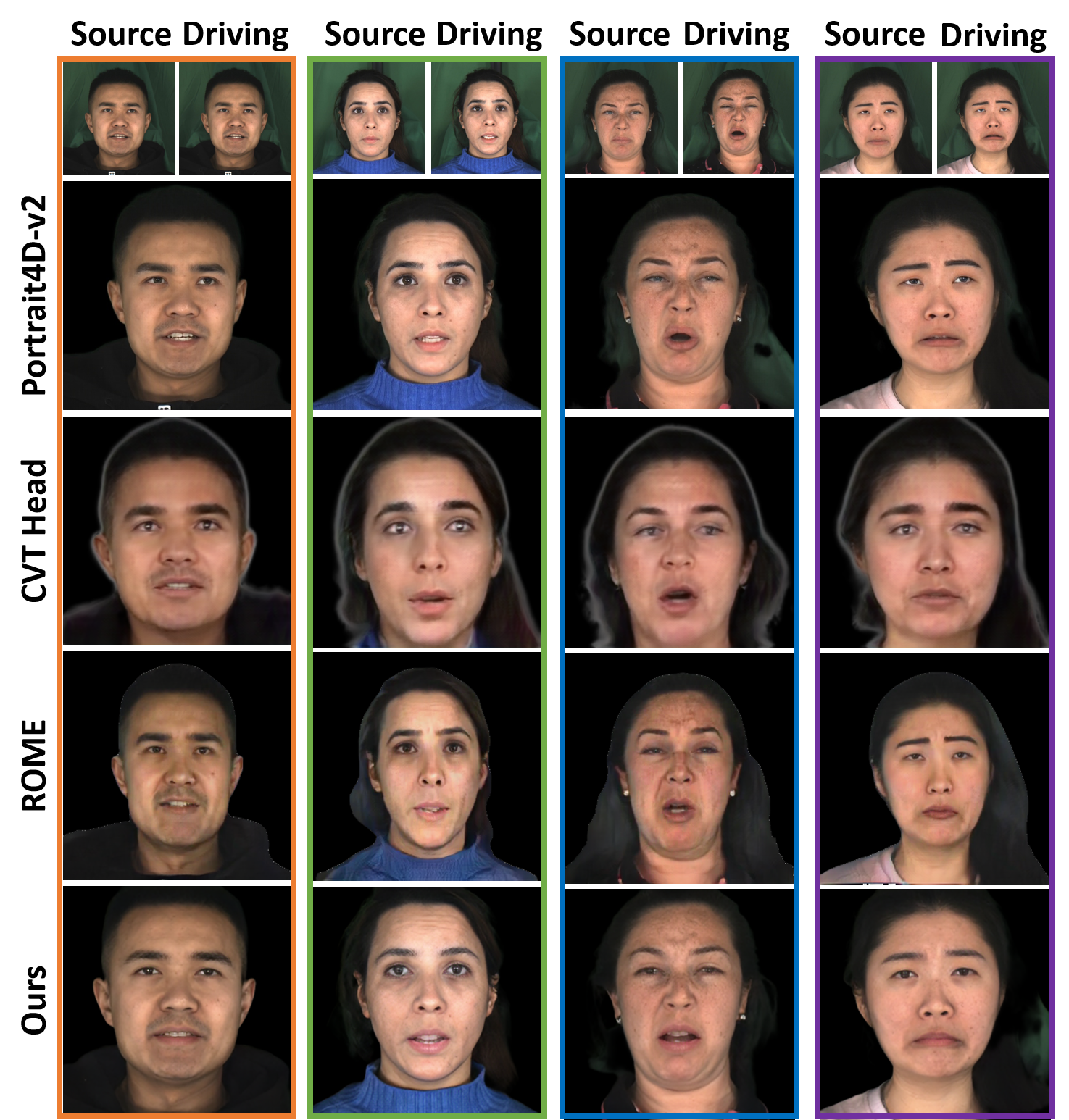}
  \caption{Qualitative comparison of one-shot full head avatar generation.}
  \label{fig:avatarRes24}
\end{figure}

\paragraph{3D Head Avatar}
For 3D avatar generation, we compare rendered images with ground truth images. Our model shows significant improvements in key metrics, achieving the highest SSIM and PSNR, along with the lowest LPIPS, as shown in Table \ref{tab:OneShotAvatar}. It indicates that our model excels in maintaining both structural integrity and perceptual quality. The high SSIM reflects our model's ability to accurately capture fine details and facial features, while the superior PSNR highlights its robustness in minimizing reconstruction noise and artifacts. Moreover, the lower LPIPS  suggests that our method produces image reconstructions that are perceptually closer to the ground truth, ensuring high-fidelity 3D avatars with realistic texture. Although Portrait4D-v2 performs competitively and ranks slightly behind our model, the noticeable jittering in the output faces during eye blinks affects the overall realism. ROME and CVTHead exhibit more challenges in preserving facial identity, reflected in higher LPIPS and lower SSIM. CVTHead, in particular, struggles to maintain identity consistency across different poses, as shown in Figure \ref{fig:avatarRes24}. In contrast, our approach preserves identity more effectively, yielding visually accurate avatars that are faithful to the subject's original appearance. This underscores our model's ability to generate high-quality, realistic 3D avatars with improved generalization across diverse inputs.

\paragraph{Realtime Face Reconstruction}
To assess the real-time performance of our 2D face reconstruction model for reverse pass-through on VR headsets like the Apple Vision Pro (with Apple M2 chip) and Meta Quest 3, we tested the model and compared inference times on the A100 GPU (used for training) with the Apple M2 SoC (in the Vision Pro), tested on a MacBook Air (8-core CPU, 4 performance cores, 4 efficiency cores, 10-core GPU, 16-core Neural Engine, and 16GB unified memory), as well as on the NVIDIA MX350 and GTX1050. Table \ref{tab:face_eyes_reconstruction_inference_m2_a100_mx350} shows that the Apple M2, with MPS acceleration, delivers inference times comparable to the A100, demonstrating its ability to efficiently run complex models in real-time. The NVIDIA MX350, though slower due to its older architecture, and the GTX1050, showed promising performance for model inference.

\input{tables/ablation}
\paragraph{Ablation Study}
Table \ref{tab:ablation} presents the quantitative results of ablation experiments. \textbf{Eye alignment ablation:} We compare the performance of Cycle-GAN versus Auto Encoder (AE) \cite{bank2021autoencoders} for eye and face alignment, as described in section \ref{subsec:Eye and Face Alignment}. The results show that Cycle-GAN outperforms AE in alignment tasks, leading to better reconstruction quality. \textbf{Cross attention ablation:} The model's performance significantly degrades when the reference and input features are concatenated, rather than using cross attention. This highlights the importance of cross attention in capturing fine-grained details for accurate reconstruction. \textbf{Reference image ablation:} Omitting the reference image results in lower performance, as it limits the model's ability to accurately reconstruct occluded areas, which are critical for realistic face restoration. The absence of this contextual information hinders the model's ability to recover facial features that are obstructed in the input. \textbf{LPIPS ablation:} Excluding LPIPS loss degrades the perceptual quality of the generated images, as evidenced by increased LPIPS score. Including it helps the model generate more visually accurate and perceptually consistent reconstructions by optimizing for human perception rather than pixel-wise similarity alone.

\section{Conclusion}
We introduce RevAvatar, an AI-driven solution to mitigate social isolation induced by VR headsets by restoring full-face images from tracking cameras using a user’s DP image, enabling real-time eye movement display on an outward-facing VR screen. Additionally, RevAvatar generates realistic one-shot full-head avatars for VR meetings and interactions. As AR/VR continues to revolutionize digital interaction, we support this advancement with VR-Face, a dataset designed to simulate real-world VR scenarios and drive research in this field. Through RevAvatar and VR-Face, we aim to set new benchmarks for AI-driven VR experiences, enhancing social presence and immersion.

%% file: tables/avatar.tex
\begin{table}[!htp]
    \centering\fontsize{9}{9}\selectfont
    \begin{tabular}{cccc}
        \toprule
        \textbf{Model}  &\textbf{SSIM}$\uparrow$ & \textbf{PSNR}$\uparrow$ & \textbf{LPIPS}$\downarrow$ \\
        \midrule
        ROME (ECCV'22) & {0.7522} & {22.7538} & {0.1089} \\
        CVTHead (WACV'24) & {0.7616} & {21.5395} & {0.1368} \\
        Portrait4D-v2 (ECCV'24) & 0.7922 & 24.3271 & 0.0638 \\
        Ours  &  \textbf{0.8025} & \textbf{25.1284} & \textbf{0.0629} \\
        \bottomrule
    \end{tabular}
    \caption{Comparison of one-shot avatar models. }
    \label{tab:OneShotAvatar}
\end{table}

%% file: tables/ablation.tex
\begin{table}[!htp]
    \centering\fontsize{9}{9}\selectfont
    \begin{tabular}{cccc}
        \toprule
        \textbf{Model}  &\textbf{SSIM}$\uparrow$ & \textbf{PSNR}$\uparrow$ & \textbf{LPIPS}$\downarrow$ \\
        \midrule
        
        2D Face-Recon (original) & \textbf{0.9445} & \textbf{31.395} & \textbf{0.0243}   \\
        AE alignment model & 0.8725  & 27.2211 & 0.1025  \\
        w/o cross attention & 0.9124 & 27.4019 & 0.0921 \\
        w/o LPIPS loss & 0.9102 & 29.0121 & 0.1001  \\
        w/o Reference Image & 0.8921 & 28.209 & 0.1129  \\
        
        \bottomrule
    \end{tabular}
    \caption{Ablation study for framework components.}
    \label{tab:ablation}
\end{table}